\documentclass[10pt,letterpaper]{article}
\usepackage[top=0.85in,left=2.75in,footskip=0.75in,marginparwidth=2in]{geometry}

\usepackage[utf8]{inputenc}

\usepackage{cite}
\usepackage{amsmath}
\usepackage{algorithm,algpseudocode}

\usepackage{nameref,hyperref}

\usepackage{microtype}
\DisableLigatures[f]{encoding = *, family = * }

\raggedright
\usepackage{changepage}

\usepackage[aboveskip=1pt,labelfont=bf,labelsep=period,singlelinecheck=off]{caption}

\makeatletter
\renewcommand{\@biblabel}[1]{\quad#1.}
\makeatother

\usepackage{lastpage,fancyhdr,graphicx}
\usepackage{epstopdf}
\fancyheadoffset[L]{2.25in}
\fancyfootoffset[L]{2.25in}

\usepackage{color}

\definecolor{Gray}{gray}{.25}

\usepackage{graphicx}

\usepackage{sidecap}

\usepackage{wrapfig}
\usepackage[pscoord]{eso-pic}
\usepackage[fulladjust]{marginnote}
\reversemarginpar

\begin{document}
\vspace*{0.35in}

\begin{flushleft}
{\Large
\textbf\newline{On-line collision avoidance for collaborative robot manipulators by adjusting off-line generated paths: an industrial use case}
}
\newline
\\
Mohammad Safeea\textsuperscript{1},
Pedro Neto\textsuperscript{2,*},
Richard Bearee\textsuperscript{1}
\\
\bigskip
\bf{1} University of Coimbra, Department of Mechanical Engineering, 3030-788 Coimbra, Portugal
\\
\bf{2} Arts et Métiers, LISPEN, 59800 Lille, France
\\
\bigskip
* pedro.neto@dem.uc.pt

\end{flushleft}

\section*{Abstract}
\textcolor{black}{Human-robot collision avoidance is a key in collaborative
robotics and in the framework of Industry 4.0. It plays an important
role for achieving safety criteria while having humans and machines
working side-by-side in unstructured and time-varying environment.
This study introduces the subject of manipulator’s on-line collision
avoidance into a real industrial application implementing typical
sensors and a commonly used collaborative industrial manipulator,
KUKA iiwa. In the proposed methodology, the human co-worker and the
robot are represented by geometric primitives (capsules). The minimum
distance and relative velocity between them is calculated, when human/obstacles
are nearby the concept of hypothetical repulsion and attraction vectors
is used. By coupling this concept with a mathematical representation
of robot’s kinematics, a task level control with collision avoidance
capability is achieved. Consequently, the off-line generated nominal
path of the industrial task is modified on–the-fly so the robot is
able to avoid collision with the co-worker safely while being able
to fulfill the industrial operation. To guarantee motion continuity
when switching between different tasks, the notion of repulsion-vector-reshaping
is introduced. Tests on an assembly robotic cell in automotive industry
show that the robot moves smoothly and avoids collisions successfully
by adjusting the off-line generated nominal paths.}

\section{Introduction}
Industrial robots are traditionally working inside fences, isolated
from humans. The ability to have robots sharing the workspace and
working side-by-side with human co-workers is a key factor for the
materialization of the Industry 4.0 concept. The paradigm for robot
usage has changed in the last few years, from an idea in which robots
work with complete autonomy to a scenario where robots cognitively
collaborate with human beings. This brings together the best of each
partner, robot and human, by combining coordination, dexterity and
cognitive capabilities of humans with the robots’ accuracy, agility
and ability to produce repetitive work\textcolor{black}{. For example,
robots can help humans in carrying and manipulating sensitive/heavy
objects safely \cite{Solanes2018} and positioning them precisely
by hand-guiding \cite{Safeea2017}. In this scenario the robot can
play the role of a force magnifier while moving compliantly according
to the haptic feedback from the human.}

\textcolor{black}{Reaching the goal of developing/creating safe collaborative
robots will allow a greater presence of robots in our society, with
a positive impact in several domains, including industry \cite{Neto2018}.
Nowadays, industrial collaborative robots, which are not operating
inside fences, do not have autonomy to perceive its unstructured and
time-varying surrounding environment, nor the ability to} avoid collisions
with human co-workers in \textcolor{black}{r}eal-time while keeping
the task target \textcolor{black}{defined by the off-line generated
paths}. On the contrary, they stop when a predefined minimum separation
distance is reached. Due to this issue, the full potentialities of
collaborative robots in industrial environment are not totally explored.
The increasing demand by industry for collaborative robot-based solutions
makes the need for advanced collision avoidance strategies more visible.
\textcolor{black}{To have them working safely alongside with humans,
robots need to be provided with biological-like reflexes, allowing
them to circumvent obstacles and avoid collisions. This is extremely
important in order to give robots more autonomy and minimum need for
human intervention, especially when robots are operating in dynamic
environment and interacting/collaborating with human co-workers \cite{Nikolaidis2015}.
}The requirements for safe collaborative robots, including physical
human-robot interaction (pHRI), are detailed in \cite{Haddadin2009},
where collision avoidance is listed as a factor, among others, which
is important for human-robot safety. The new standard ISO 10218 and
the technical specification TS 15066 define the safety requirements
for collaborative robots \cite{saenz2018}. Apart from industrial
domain and human-robot collaboration, collision avoidance is also
being investigated for aerospace applications, including robotic arms
mounted on space maneuverable platforms \cite{Chu2018} and aerial
manipulators mounted on drones \cite{Jeon2017}.

\textcolor{black}{For a proper implementation of collision avoidance
in Human-Robot Interaction (HRI) scenario, the motion of the human
co-worker shall be predicted, and his/her configuration shall be captured.
For this purpose, researches have utilized various methods and sensors.
In \cite{5152690}, the authors presented a method for human pose
estimation (in a plane) based on laser range measurements, the method
was applied successfully in HRI scenario between a human and a mobile
robot. The intelligent space (iSpace) project was introduced in \cite{1511628},
iSpace implements a distributed network of sensors for tracking the
motion of humans, the iSpace provides a platform for guiding mobile
robots in human cantered environment. In \cite{Flacco2012} the authors
used the depth data from a Kinect camera for calculating distances
between the human and reference points on the robot. In \cite{8206225}
the authors presented a new method, by using a camera mounted on the
EEF (eye-in-hand) or the worker’s head, a coordinated motion between
a robot and a human was achieved, for future work the authors plan
to extend the method for achieving human-robot collision avoidance.}

\textcolor{black}{After capturing the motion of the human(s)/obstacle(s),
the collision avoidance shall be developed. In the robotics literature
various studies have been proposed. In the pioneering work of Khatib
\cite{Khatib1986}, a real-time obstacle avoidance approach based
on the classical artificial potential field (PF) concept is introduced.
In PF-based methods, the robot is in a hypothetical vector field,
and its motion is influenced by forces of attraction that guide the
robot towards the goal and forces of repulsion that repel it away
from obstacles. Subjected to these forces the robot finds its way
to the goal while avoiding collisions. Recently, a depth space approach
for collision avoidance proposes an improved implementation of the
PF method in which an estimation of obstacle’s velocity was taken
into consideration when computing the repulsion vector \cite{7421962}.
PF-based robot self-collision avoidance has been studied, as well
as the development of collision avoidance techniques for redundant
robots \cite{7097068}. A distributed real-time approach to collision
avoidance considering not only the robot tool centre point over the
objects in the cell, but also the body of the tool mounted on the
robot flange is in \cite{7005160}. A passivity-based control scheme
for human-robot safe cooperation is proposed in \cite{ZanchettinTAND}.
A collision free trajectory generating method for a robot operating
in a shared workspace in which a neural network is applied to create
the way points required for dynamic obstacles avoidance is proposed
in \cite{MEZIANE2017243}. In \cite{Rubio2016}, the authors presented
a method for calculating collision free optimal trajectories for robotic
manipulators with static obstacles. The proposed algorithm takes into
consideration the maximum limits of jerk, torques, and power for each
actuator. Tests have been carried out in simulation in a PUMA 560
robot. In \cite{Perdereau2002}, it is presented a collision avoidance
algorithm between robotic manipulators and mobile obstacles validated
in simulation environment. Using the variation principal, it is proposed
a path planner for serial manipulators with high degrees of freedom
operating in constrained work spaces where the planner produces monotonically
optimal collision free paths \cite{Shukla2013}. Based on fuzzy rules
\cite{PALM1992279}, the authors presented a method for resolving
internal joint angles in redundant manipulators. The method allows
the EEF to follow the desired path, while the internal motion manifold
is used to perform other objectives including collision avoidance
with surrounding obstacles.}

While artificial PF is inspired by electric field phenomena, other
approaches, inspired by electromagnetism (circular fields) were investigated
\cite{HADDADIN20116842,Singh1996}. Other researchers have approached
robot collision avoidance using optimization techniques \cite{Bosscher2011},
by formulating the problem as a graph search using Probabilistic Roadmaps
(PRM) \cite{Kavraki1996}, and considering dynamic changing environment
\cite{Leven2002}. However, the number of existing studies dedicated
to on-line human-robot collision avoidance for manipulators in industrial
setups is very limited, and when existing, results are presented in
simulation environment. Some of these studies, especially the ones
with more direct industrial application, approach collision avoidance
by stopping the robot or reducing its velocity when a human reaches
a given distance threshold \cite{Schmidt2014711}. An interesting
work defines four safety strategies for workspace monitoring and collision
detection: the system alerts the operator, stops the robot, moves
the robot away, or modifies the robot’s trajectory from an approaching
operator \textcolor{black}{\cite{wang2017}}.

\begin{figure}[ht]
    \centering
    \includegraphics[width=0.9\textwidth]{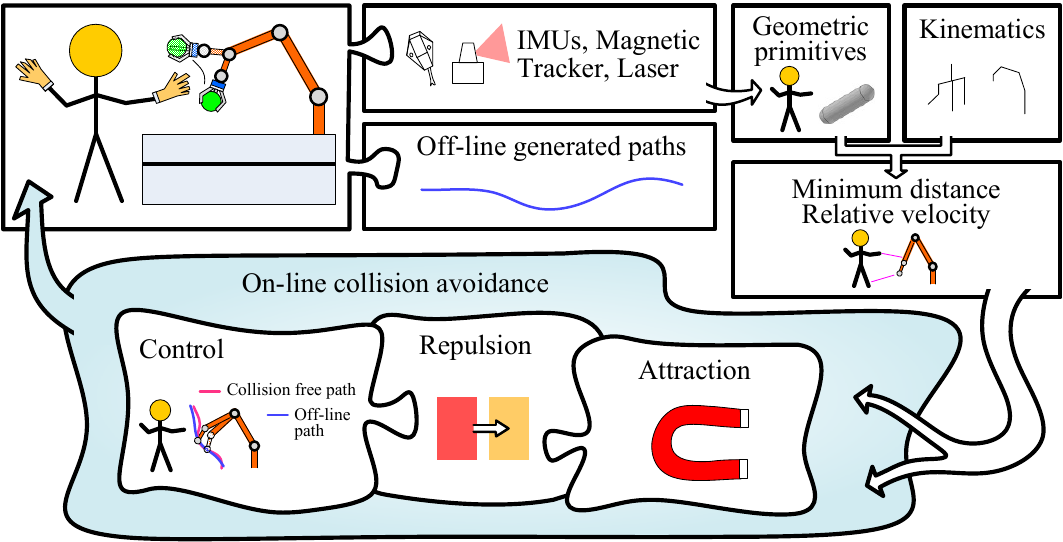}
    \caption{Proposed framework for on-line human-robot collision avoidance.}
    \label{fig:fig1}
\end{figure}

\textcolor{black}{In our study, an industrial task is defined by the
off-line generated robot paths (task primitives), which can be programmed
by a Computer-Aided Design (CAD) software, or by using more sophisticated
methods including Programming by Demonstration (PbD) \cite{skoglund2007programming}.
Consequently, a safe human robot interaction is achieved through real-time
modification of the off-line generated paths, Fig. \ref{fig:fig1}.
}In this scenario of shared workspace, the human co-worker focuses
on the collaborative task he/she is performing rather than the potential
danger from the robot. Using external sensors (IMUs, a magnetic tracker
and a laser scanner) the pose of the human body is captured and approximated
by hemisphere-capped cylinders designated in this paper by capsules.
The robot is represented by three capsules. The analytical minimum
distance between capsules representing the robot and the human(s)
is calculated using the method in \cite{8586786}. The human-robot
minimum distance and relative velocity are used as inputs for the
proposed collision avoidance controller, where hypothetical attraction
and repulsion vectors attract the robot towards the goal/target while
repelling it away from obstacles. We also introduce the notion of
repulsion-vector-reshaping to avoid control discontinuity. By coupling
these concepts with a mathematical representation of robot’s kinematics
we can achieve a task level control with smooth collision avoidance
capability. The proposed framework is tested in three different configurations,
including a real use case for assembly in automotive industry using
real sensors and a collaborative industrial manipulator. Results indicate
that the robot reacts smoothly by modifying its off-line generated
paths to avoid collision with the human co-worker.\textcolor{black}{{}
Consequently, our study is the first of its kind (according to our
knowledge) that satisfies all the following points combined:}
\begin{enumerate}
\item \textcolor{black}{In our study a real industrial robot (not experimental)
is used for performing a typical industrial task;}
\item \textcolor{black}{The proposed method for performing the collision
avoidance is tailored for industrial use, by combining an off-line
path of the EEF (important for industrial applications) with an on-line
reactive collision avoidance (required for dynamic collision avoidance);}
\item \textcolor{black}{In our method a human co-worker is moving freely
around the robot, the whole configuration of his upper body is captured
using real sensors. While, most other studies utilize simulation,
the few that approached human-robot collision avoidance in a real
scenario utilize an experimental robot, and mostly vision sensing
which suffers from occlusion. In addition, other studies focused on
the collision avoidance itself, without showing results in a real
industrial operation;}
\item \textcolor{black}{Unlike other studies, we realized that in a real
industrial scenario, the control shall switch between different operation
modes (as shown later in Fig. \ref{fig:fig11} and Algorithm
Collision avoidance – automotive sample case), leading to repulsion
action discontinuity. We solved this issue by proposing the repulsion-vector-reshaping
(described in Algorithm Modified repulsion vector).}
\end{enumerate}

\section{Challenges and problem Specification\label{subsec:Problem-Specification}}

Two major problems in on-line human-robot collision avoidance can
be identified. The first is related with the reliable acquisition
of the human pose in unstructured environments. The second is due
to the difficulty in achieving smooth continuous robot motion while
generating collision avoidance paths. For capturing the configuration
of the human the method in \cite{Mohammad2018} is implemented. On
the other hand, our study presents solutions concerning the difficulty
in achieving collision free and smooth continuous robot motion, which
is particularly visible in an industrial setup where the control algorithm
switches between different controllers depending on the task-in-hand.
In summary, several challenges can be pointed out:
\begin{enumerate}
\item Accurate definition of humans/obstacles and robot pose in space using
geometric primitives, and calculation of the minimum distance between
them;
\item Achieving reliable autonomous human-robot collision avoidance in which
the robot adapts the off-line generated nominal paths while keeping
the task goal/target. In such a case, instead of stopping or reducing
robot's velocity when humans are nearby, the robot has to continue
its motion while avoiding the humans/obstacles;
\item The control strategy shall produce continuous motion of robot's reaction
when it adjusts the path to avoid collision. This continuity shall
be guaranteed even when switching between different controllers;
\item Industrial applications require high-performance control in terms
of motion accuracy and agility;
\item Collision free robot motion should be possible and reliable in the
entire working volume of the robot.
\end{enumerate}
Experiments demonstrated the ability of the proposed solution to achieve
on-line human-robot collision avoidance materialized in the following
contributions:
\begin{enumerate}
\item Reliable and smooth human-robot collision avoidance in which the robot
adapts the off-line generated nominal paths (defined in the initial
robot program) while keeping the task target. The robot finds a way
to get around the human(s)/obstacles when they are nearby. Human-robot
minimum distance and relative velocity are used as inputs to the implemented
collision avoidance algorithm;
\item Successfully applying on-line collision avoidance on a real industrial
collaborative robot performing industrial assembly tasks in collaboration
with a human co-worker.
\end{enumerate}

\section{Collision Avoidance Strategy\label{sec:Gesture-Segmentation}}

Hypothetical attraction and repulsion vectors attract the robot towards
the goal/target (defined by the off-line generated nominal paths)
while repelling it away from human(s)/obstacles. By coupling this
concept with a mathematical representation of robot's kinematics we
can achieve a task level control with collision avoidance capability.

\subsection{Repulsion}

A vector $\boldsymbol{v_{\mathrm{\mathit{cp.rep}}}}$ acts on the
point of the robot closest to the obstacle (CP) repelling it away
from collision. This vector is defined considering a magnitude $v_{rep.mod}$
(calculated from a base repulsion amplitude $v_{rep}$) and a direction
$\boldsymbol{s}$:

\begin{equation}
\boldsymbol{v_{cp.rep}}=v_{rep.mod}\boldsymbol{s}\label{eq:repulsion_vector}
\end{equation}

\noindent The direction of the repulsion vector $\boldsymbol{s}$
is taken to be aligned with the line segment associated with the minimum
distance:

\begin{equation}
\boldsymbol{s}=\frac{\boldsymbol{r_{1}-r_{2}}}{|\boldsymbol{r_{1}-r_{2}}|}
\end{equation}

\noindent Where $\boldsymbol{r_{1}}$ is the position vector of the
point of the robot closest to the obstacle and $\boldsymbol{r_{2}}$
is the position vector of the point of the obstacle closest to the
robot. For calculating the base repulsion amplitude $v_{rep}$ we
propose to superimpose the repulsion due to the minimum distance ($v_{rep1}$)
and the repulsion due to the relative velocity between the human and
the robot ($v_{rep2}$), so that $v_{rep}=v_{rep1}+v_{rep2}$. Here,
$v_{rep1}$ is calculated from the minimum distance $d_{min}$:

\begin{equation}
v_{rep1}=\begin{cases}
k_{1}\left(\frac{d_{0}}{d_{min}-d_{cr}}-1\right), & \:if\;d_{min}-d_{cr}<d_{0}\\
0 & \:if\;d_{min}-d_{cr}\geq d_{0}
\end{cases}\label{eq:repulsion_force_mag}
\end{equation}

\noindent Where $k_{1}$ is a constant, \textcolor{black}{$d_{0}$
is an offset distance around the obstacle’s capsule, it specifies
the area around the obstacle where the repulsion vector is activated,
}and $d_{cr}$ is a critical distance below which the robot is not
allowed to be near the human. To enhance the responsiveness of the
robot, we propose a dynamical reshaping of the size of the area of
influence around the obstacle $d_{0}$, such that the value of $d_{0}$
increases when the relative velocity between the robot and the obstacle
increases:

\begin{equation}
d_{0}=\begin{cases}
d_{1}-c_{v}v_{rel} & v_{rel}<0\\
d_{1} & v_{rel}\geq0
\end{cases}
\end{equation}

\noindent Where $v_{rel}$ is the human-robot relative velocity, $c_{v}$
is a constant and $d_{1}$ is the minimum value of the area of influence
around the obstacle. For $v_{rep2}$ we have:

\begin{equation}
v_{rep2}=\begin{cases}
-c\,k_{2}\,v_{rel} & v_{rel}<0\\
0 & v_{rel}\geq0
\end{cases}
\end{equation}

\noindent Where $k_{2}$ is a damping constant and $c$ is a coefficient
that takes into consideration the proximity of the obstacle from the
robot:

\begin{equation}
c=\begin{cases}
1 & d_{min}<l_{1}\\
\frac{1+cos(\pi\frac{d_{min}-l_{1}}{l_{2}-l_{1}})}{2} & l_{1}<d_{min}<l_{2}\\
0 & l_{2}<d_{min}
\end{cases}
\end{equation}

\noindent Where $l_{1}$ and $l_{2}$ are constant distances that
define the range around the robot where the damping force is activated.
The intuition of using $c$ is that obstacles far away from the robot
shall not affect robot's motion since that they do not pose any risk
of collision.

The modified repulsion magnitude $v_{rep.mod}$ is calculated from
$v_{rep}$ according to Algorithm “Modified repulsion vector”. For
complex industrial collaborative operations, the collision avoidance
controller is typically embedded in a state machine, where the collision
avoidance functionality is activated/deactivated according to the
tasks being performed. In such a case, discontinuity could appear
when calculating the repulsion action. As an example, if the controller
is switched (from the collision avoidance deactivation to the collision
avoidance activation) while the co-worker is near the robot, discontinuity
appears. In such a case, a high magnitude of the repulsion vector
will act on the robot suddenly. To solve this problem, the repulsion
action is proposed to be time dependent, by introducing the concept
of repulsion-vector-reshaping coefficient $\gamma$, such that when
the control scheme is switched the repulsion magnitude is allowed
to increase monotonically starting from zero up to its stable value.
In the Algorithm, $now$ is a function returning the current time,
$\tau$ is a time constant that can be calculated from $v_{max}/(5a_{max})$,
where $v_{max}$ and $a_{max}$ are the maximum curvilinear velocity
and acceleration of the end-effector (EEF) used during collision avoidance,
respectively.

\begin{figure}[ht]
    \centering
    \includegraphics[width=0.8\textwidth]{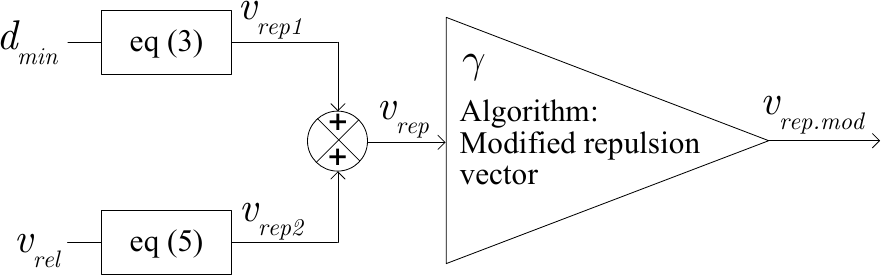}
    \caption{Block diagram showing the proposed method for calculating
the magnitude of the modified repulsion vector.}
    \label{fig:fig2}
\end{figure}

\textcolor{black}{Figure \ref{fig:fig2} shows
a block diagram illustrating the proposed method for calculating the
modified repulsion vector $v_{rep.mod}$ and its relationship to $v_{rep1}$
and $v_{rep2}$.}

$ $
\begin{algorithm}
\caption{Modified repulsion vector}
\begin{algorithmic}[1]

\For{\textbf{each} time step $\bigtriangleup t$} 
\If{controller switched}
	\State $t_0=now$
\EndIf
\State $t=now-t_0$
\State $\gamma=1-exp(-t/\tau)$
\State ${v}_{rep.mod}=\gamma {v}_{rep}$
\EndFor
\end{algorithmic}
\end{algorithm}

\begin{figure}[ht]
    \centering
    \includegraphics[width=0.75\textwidth]{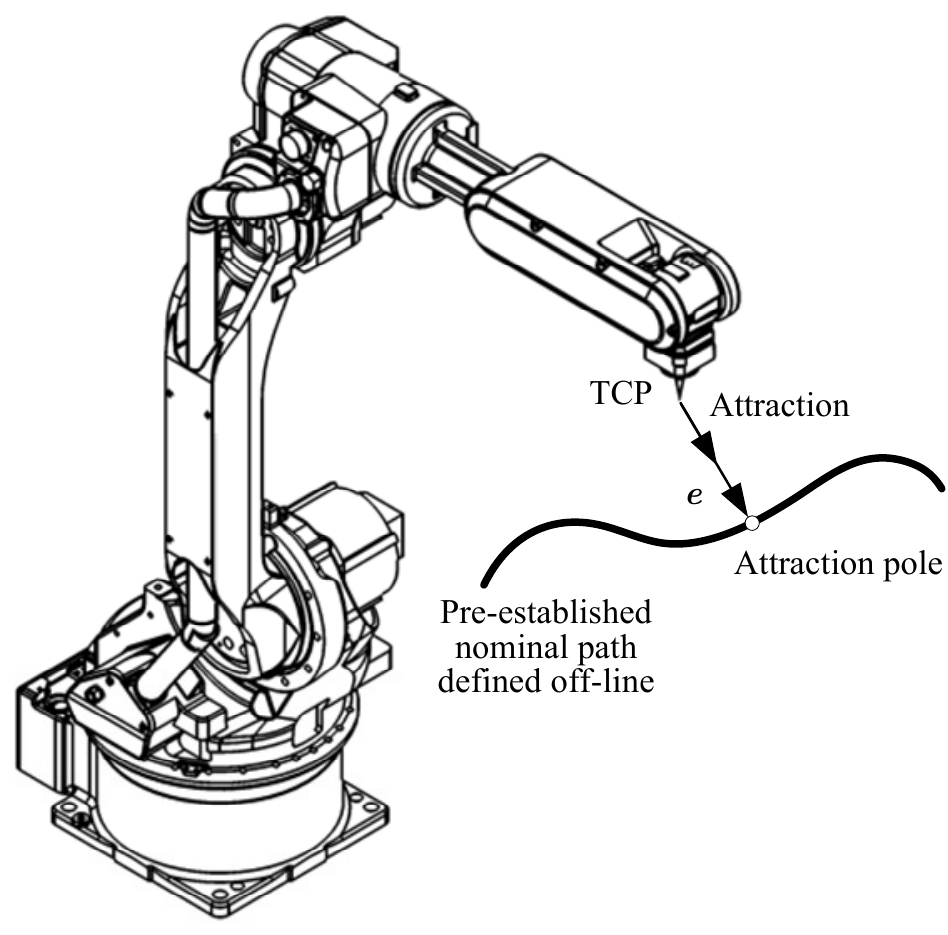}
    \caption{The nominal path curve defined off-line, the attraction pole, and the error vector.}
    \label{fig:fig3}
\end{figure}

\subsection{Attraction\label{subsec:Threshold-Optimization}}

An attraction velocity vector $\boldsymbol{v_{e.att}}$ attached to
the EEF guides the robot towards the goal/target, Fig. \ref{fig:fig3}.
This vector is a function of the error $\boldsymbol{e}$ between EEF's
position $\boldsymbol{p}_{e}$ and the goal position $\boldsymbol{p_{g}}$
(defined in the off-line generated nominal paths):
\begin{equation}
\boldsymbol{e}=\boldsymbol{p_{e}}-\boldsymbol{p_{g}}
\end{equation}
The attraction velocity is calculated from a proportional term $(\boldsymbol{\psi}_{p})$
and a quasi-integral term $(\boldsymbol{\psi}_{i})$:

\begin{equation}
\boldsymbol{v_{e.att}}=\beta(\boldsymbol{\psi}_{p}+\boldsymbol{\psi}_{i})\label{eq:attraction_vector1987}
\end{equation}
Where $\boldsymbol{\psi}_{p}$ is a pure proportional term:

\begin{equation}
\boldsymbol{\psi}_{p}=-\boldsymbol{\mathrm{K}_{p}e}
\end{equation}
In which $\boldsymbol{\mathrm{K}_{p}}$ is the proportional coefficient.
The quasi-integral term $\boldsymbol{\psi}_{i}$, Algorithm “Integral
term of the attraction vector”, prevents the quasi-integral from accumulating
when the human-robot distance is less than a predefined safety distance
$d_{0}$. In the Algorithm, $d_{min}$ is the human-robot minimum
distance and the integral term is calculated numerically using a simple
Euler scheme (more sophisticated Runge-Kutta methods could also be
used). The term $\beta$ is used to reduce the magnitude of the attraction
vector. This term has the effect of detaching the EEF gradually from
the goal when the human co-worker is closer to the robot:

\begin{equation}
\beta=\left(\frac{2}{1+e^{-\left(\frac{d_{min}-d_{cr}}{d_{0}}\right)^{2}}}-1\right)
\end{equation}

$ $
\begin{algorithm}
\caption{Integral term of the attraction vector}
\begin{algorithmic}[1]

\For{\textbf{each} time step $\bigtriangleup t$} 
\If{$d_{min}$-$d_{cr}$>$d_{0}$}
	\State $\boldsymbol{\psi}_{i}=\boldsymbol{\psi}_{i}-\boldsymbol{\mathrm{K}}_{i}\int_{t}^{t+\bigtriangleup t}\boldsymbol{e} {dt}$
\Else
	\State $\boldsymbol{\psi}_{i}=\boldsymbol{\psi}_{i}$ \Comment{\color{black}to prevent windup of integral term\color{black}}
\EndIf

\EndFor
\end{algorithmic}
\end{algorithm}

\begin{figure}[ht]
    \centering
    \includegraphics[width=0.9\textwidth]{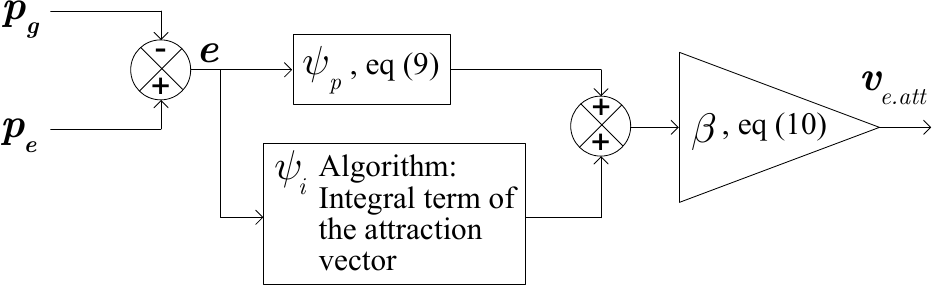}
    \caption{Block diagram showing the proposed method for calculating
the attraction vector. Term $\boldsymbol{\psi}_{i}$ is calculated
using the Algorithm Integral term of the attraction vector, which
is used to avoid windup problem.}
    \label{fig:fig4}
\end{figure}

\textcolor{black}{Figure \ref{fig:fig4} shows
a block diagram illustrating the proposed method for calculating the
attraction vector.}

\subsection{Controller}

The robot is controlled at the joint velocity level. The repulsion
and attraction vectors are considered velocity vectors in which the
repulsion velocity is calculated from (\ref{eq:repulsion_vector}),
and the attraction velocity at the EEF $\boldsymbol{v_{e.att}}$ is
calculated from (\ref{eq:attraction_vector1987}). Calculating the
overall angular velocities of the joints requires superimposing the
angular velocities due to repulsion and attraction. Thus, the angular
velocities due to $\boldsymbol{v_{cp.rep}}$ that acts at CP is calculated
using the Damped Least Squares \cite{Buss2004}:

\begin{equation}
\boldsymbol{\dot{q}_{rep}}=\boldsymbol{\mathrm{J_{cp}^{T}}}(\boldsymbol{\mathrm{J_{cp}}}\boldsymbol{\mathrm{J_{cp}^{T}}}+\lambda^{2}\mathrm{\boldsymbol{I}})^{-1}\boldsymbol{v_{cp.rep}}
\end{equation}

\noindent Where $\boldsymbol{\dot{q}_{rep}}$ is the joint velocities
vector due to the repulsion action, $\boldsymbol{\mathrm{J_{cp}}}$
is the partial Jacobian associated with CP on the robot, $\lambda$
is a damping constant, and $\mathrm{\boldsymbol{I}}$ is the identity
matrix.

\noindent The angular velocities due to $\boldsymbol{v_{e.att}}$
that act at the EEF are calculated from:

\begin{equation}
\boldsymbol{\dot{q}_{att}}=\boldsymbol{\mathrm{J_{e}^{T}}}(\boldsymbol{\mathrm{J_{e}}}\boldsymbol{\mathrm{J_{e}^{T}}}+\lambda^{2}\mathrm{\boldsymbol{I}})^{-1}\boldsymbol{v_{e.att}}
\end{equation}

\noindent Where $\boldsymbol{\dot{q}_{att}}$ is the joint velocities
vector due to the attraction action and $\boldsymbol{\mathrm{J_{e}}}$
is the Jacobian associated with the EEF. Thus, the total angular velocities
of the joints sent to the robot:

\begin{equation}
\boldsymbol{\dot{q}_{total}}=\boldsymbol{\dot{q}_{att}}+\boldsymbol{\dot{q}_{rep}}
\end{equation}

\section{Experiments and Results\label{sec:Testing-Methodology}}

Experiments are conducted in three main configurations:
\begin{enumerate}
\item Configuration 1: the human arm acts as an obstacle for the robot that
is performing a straight line path (off-line generated nominal path);
\item Configuration 2: the human approaches the robot from the side while
the robot is stopped at a predefined home position;
\item Configuration 3: an industrial collaborative assembly operation for
automotive industry in which the human co-worker approaches the robot
to place a sticker in a car door card while the robot is inserting
trim clips in the same door card.
\end{enumerate}

\subsection{Setup and Data Acquisition}

The three experimental configurations were tested using different
sensors for capturing the human pose in space. In configuration 1
and configuration 2, the proposed solution was tested with Polhemus
Liberty magnetic tracking sensors attached to the human upper body
(arm, forearm and chest) to acquire 6 DOF pose (position and orientation)
of each body part in space. In configuration 3, the method proposed
in \cite{Mohammad2018} is used for capturing the human body pose
from five IMUs (Technaid MCS) attached to the arms/forearms and the
chest, and a laser scanner (SICK TiM5xx) at the level of the legs.
An external computer Intel Core i7 with 32 GB of RAM running MATLAB\textsuperscript{\textregistered}
was used for performing the required computations: sensor data acquisition,
capsules configuration calculation, minimum-distance and relative-velocity
calculation, collision avoidance algorithms, and robot control using
the KUKA Sunrise Toolbox (KST) \cite{Safeea2018}.\textcolor{blue}{{}
}\textcolor{black}{In such a case, the robot is controlled at the
kinematics level without considering its dynamics explicitly. The
low-level control is built on the DirectServo library \cite{manualKUKA}
provided by the manufacturer (Kuka Roboter), a 25Hz angular position
update of the real-time system of the robot controller is used. Hence,
the industrial joint servo is used for controlling the joint error
dynamics, while the commanded joint position is calculated according
to the manufacturer's data in order not to exceed the maximum allowable
angular velocity. The demonstration of an efficient and fast obstacle
avoidance responsiveness is an objective of this study. However, the
joint trajectories may be smoothed by filtering \cite{BESSET2017169},
to adjust the HRI acceptance and/or to improve the dynamic accuracy.}

\begin{figure}[ht]
    \centering
    \includegraphics[width=0.75\textwidth]{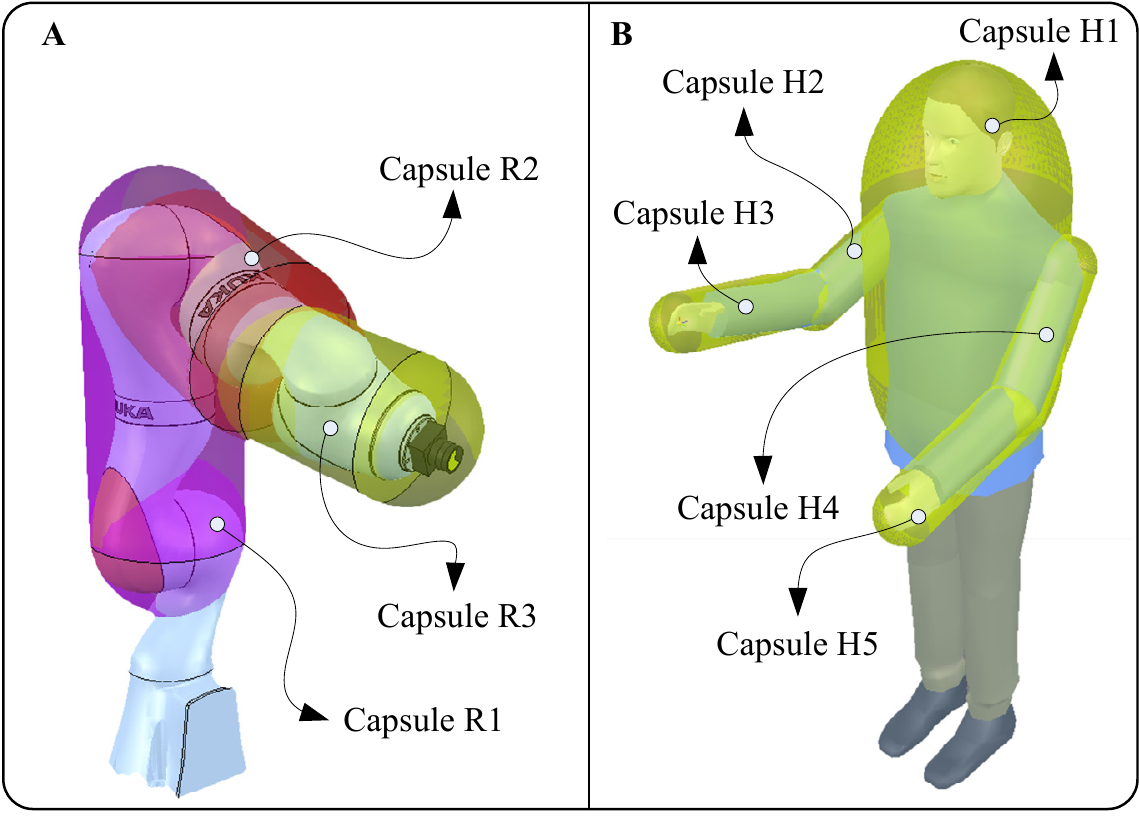}
    \caption{Robot (A) and human (B) represented by capsules.}
    \label{fig:fig5}
\end{figure}

\begin{figure}[ht]
    \centering
    \includegraphics[width=0.85\textwidth]{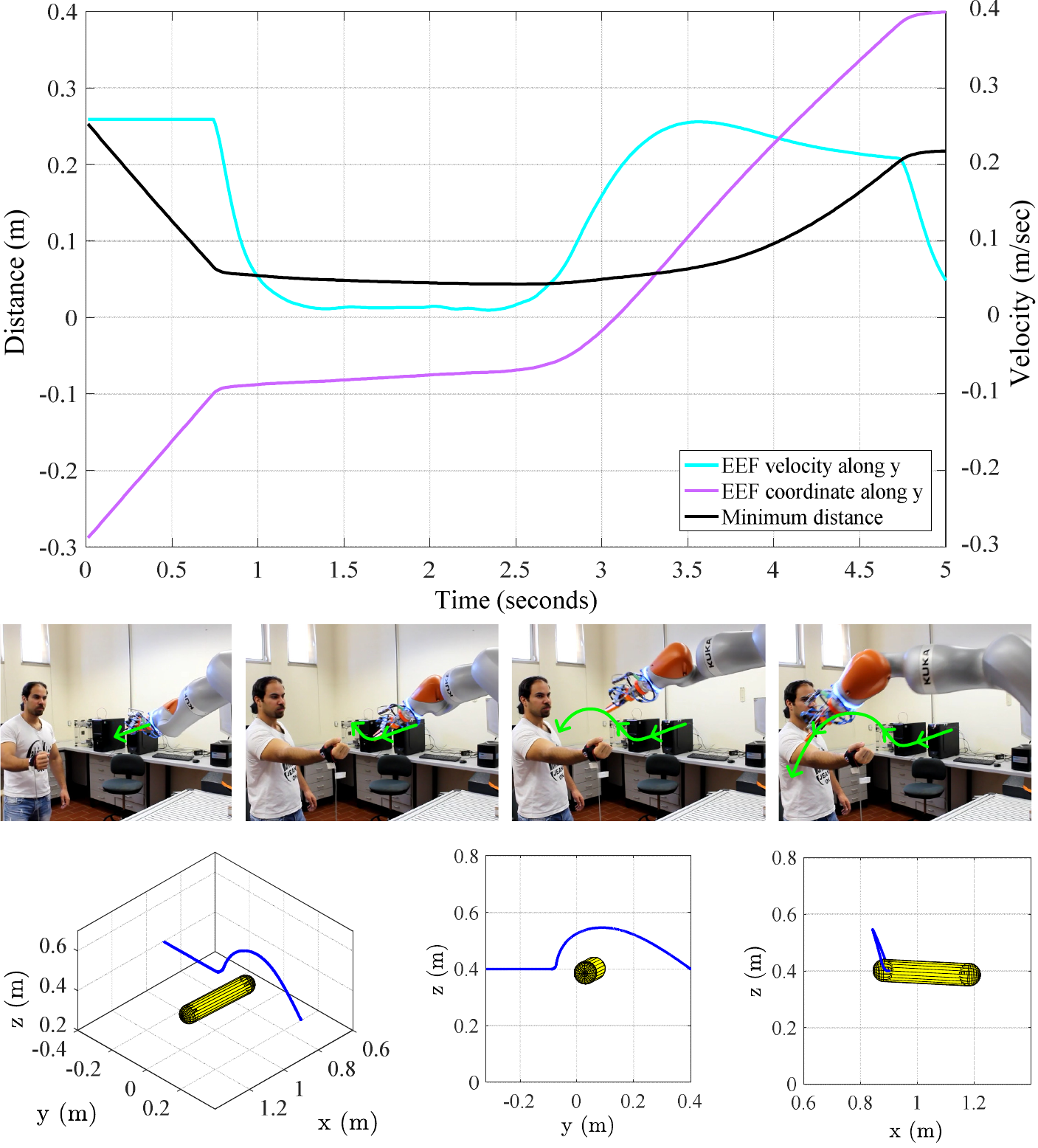}
    \caption{Configuration 1 results (first initial condition,
off-line generated path is parallel to y direction). Plot shows the
minimum distance, EEF velocity and position along y axis (top). Snapshot
of collision avoidance testing and collision avoidance path in 3D
and 2D space (middle and bottom).}
    \label{fig:fig6}
\end{figure}

\subsection{Human and Robot Representation}

The human is represented by five capsules, Fig. \ref{fig:fig5},
four capsules used to cover the right/left upper arm and forearm,
while the fifth capsule is used to cover the torso up to the head
\cite{Mohammad2018}. The robot (KUKA iiwa with 7 DOF) is represented
by three capsules, Fig. \ref{fig:fig5}.
Capsule R3 also incorporates the tool attached to the robot. The pose
of the capsules are defined by applying the forward kinematics on
the joint angles acquired from the controller of the robot.

\subsection{Results and Discussion}

\begin{figure}[ht]
    \centering
    \includegraphics[width=0.9\textwidth]{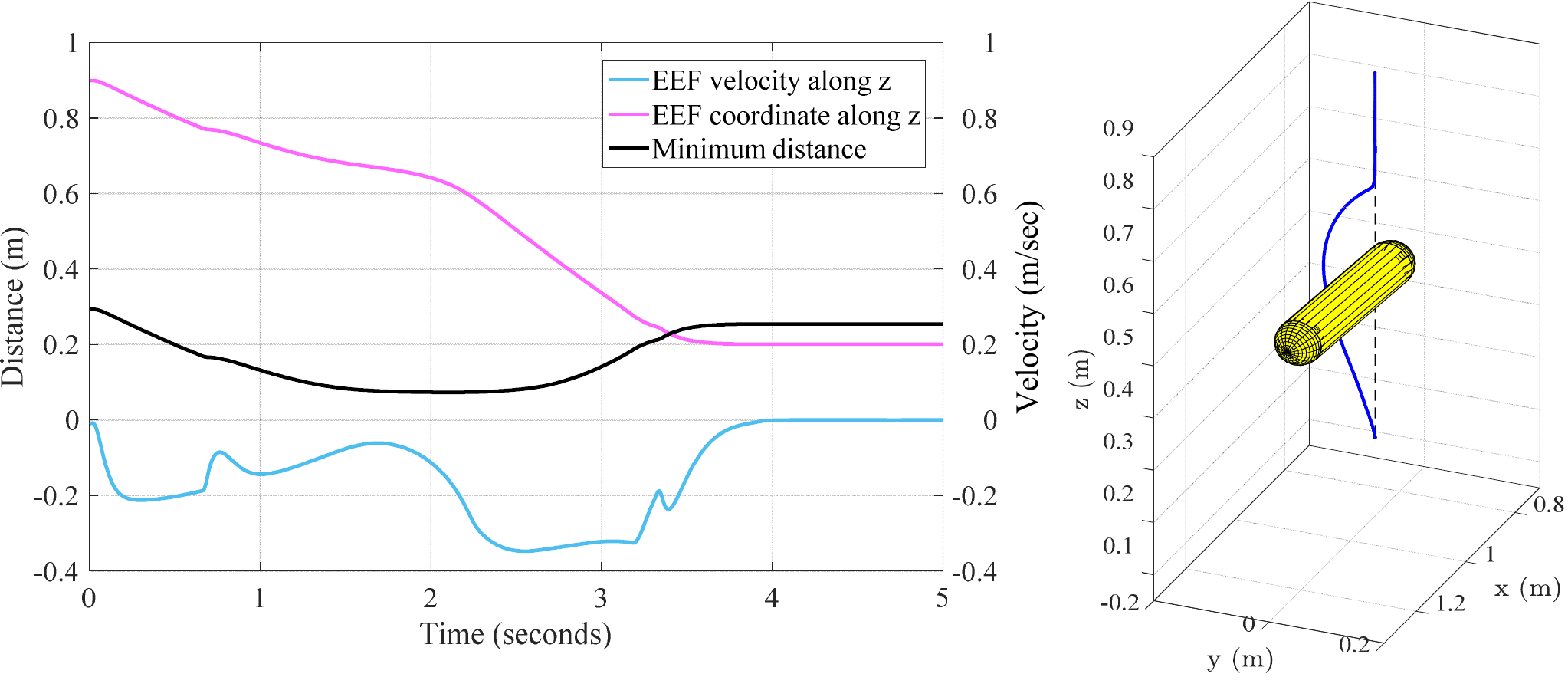}
    \caption{Configuration 1 results (second initial condition,
off-line generated path is parallel to $z$ axis). Plot shows the
minimum distance, EEF velocity and position along $z$ axis (left).
Path of EEF in 3D (right).}
    \label{fig:fig7}
\end{figure}

\begin{figure}[ht]
    \centering
    \includegraphics[width=0.8\textwidth]{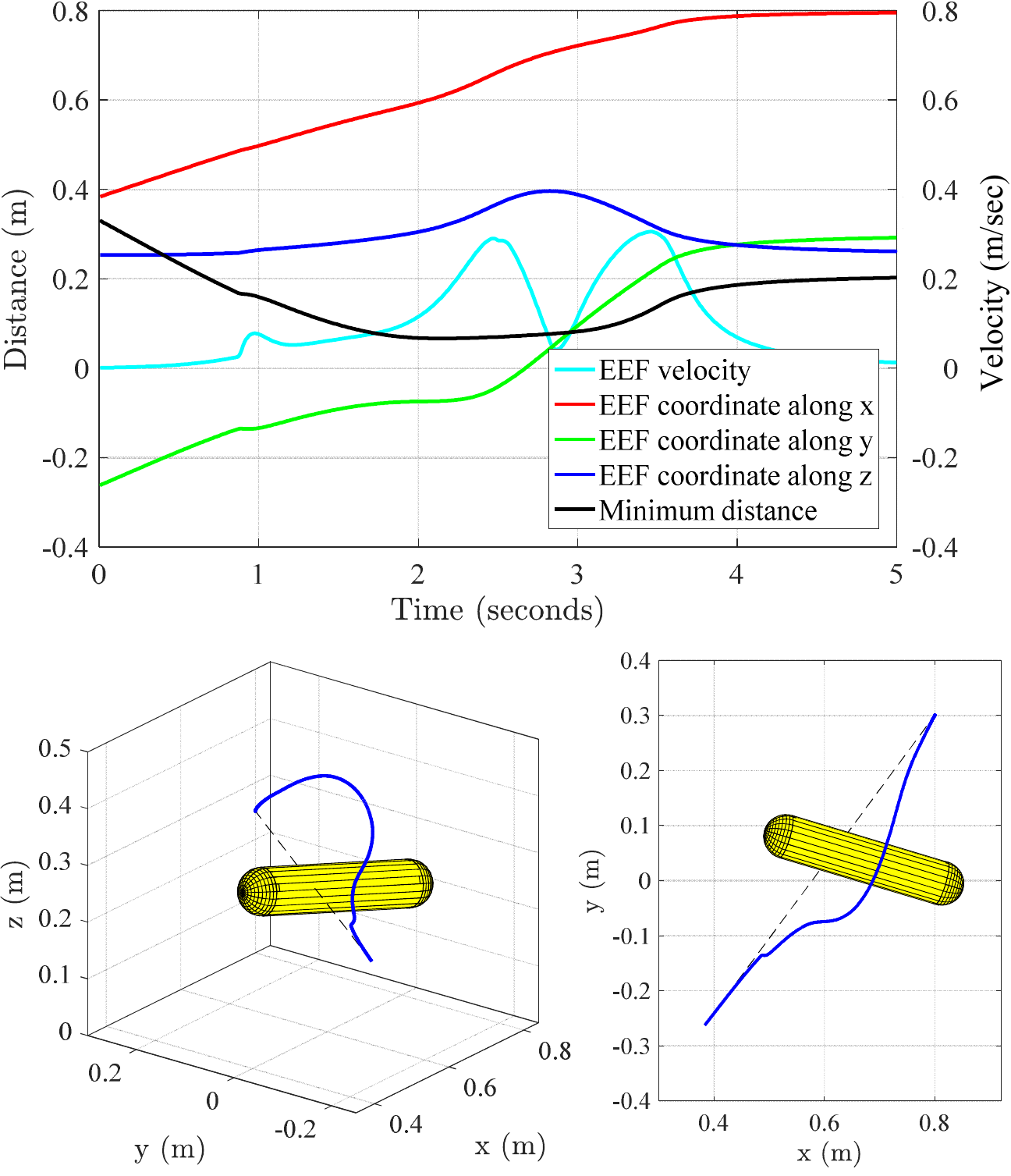}
    \caption{Configuration 1 results (third initial condition,
off-line generated path is inclined in a plane parallel to $xy$ of
the robot base). Plot shows the minimum distance, EEF velocity and
position (top). Path of EEF in 3D (bottom).}
    \label{fig:fig8}
\end{figure}

Figure \ref{fig:fig6} shows the results for configuration
1. The human forearm, represented by a capsule, is extende\textcolor{black}{d
(almost parallel to the $x$ axis of the robot base frame) }and acts
as an obstacle. The robot is moving on a straight line path (nominal
path defined off-line) along the y direction of its base frame. While
moving on the straight line, the minimum distance between the human-arm
and the robot decreases. Consequently, the robot adapts the nominal
path smoothly circumventing the human arm. At the top of Fig. \ref{fig:fig6},
the graph shows the minimum distance, the velocity of the EEF and
its position in Cartesian coordinates along y axis. These quantities
are reported as function of time. We can notice from the plot that
at the beginning the human arm is in a resting position and the robot
is moving with a constant velocity of about 0.26 m/sec, along y direction,
towards the human arm. When the robot EEF approaches the human arm
the minimum distance decreases to a minimum of about 5 centimeters.
The EEF velocity is constant until a threshold minimum distance is
reached. In such scenario the EEF velocity decreases to start circumventing
the obstacle and then accelerates to reach a velocity close to the
nominal velocity of 0.26 m/s.\textcolor{black}{{} We also tested the
system in configuration 1 but with different initial conditions through
changing the robot's initial configuration, the off-line generated
path and the position/orientation of the obstacle (human's arm). Figure
\ref{fig:fig7} shows the results
in the case where the off-line generated path is parallel to the $z$
axis, the robot is required to move on the path in the negative $z$
direction, the human forearm (almost parallel to the $x$ axis of
the robot base frame) is extended in the way of the off-line generated
path. From Fig. \ref{fig:fig7}
(right) it is noticed that the robot adapts the off-line path successfully
using the proposed method. Figure \ref{fig:fig7}
(left) shows the minimum distance, the velocity and the position of
the EEF along the $z$ axis (following the robot motion from $z=0.9$
m at top to $z=0.2$ m at the bottom), the minimum distance reached
is around 7 cm, the velocity profile differs a little bit from the
previous case (motion along the $y$ axis), but the general behaviour
is the same. In Fig. \ref{fig:fig8} the same
test (configuration 1) was repeated with yet another initial condition,
the obstacle (human arm) is inclined, and the off-line generated path
is parallel to the $xy$ plane of the robot base (starting from point
$x=0.38$, $y=-0.26$, $z=0.25$ to the point $x=0.8$, $y=0.3$,
$z=0.25$). Figure \ref{fig:fig8} at the top
shows the coordinates of the EEF, the minimum distance and the magnitude
of EEF's velocity acquired during the collision avoidance motion,
it can be noticed that the minimum distance reached is around 6.5
cm. Figure \ref{fig:fig8} at bottom shows the
obstacle and robot's path in perspective view (left) and in the $xy$
plane (right). From previous results it can b}e concluded that the
robot manages to avoid collision with the co-worker successfully and
the collision avoidance controller smoothly reacts to avoid collision
while reaching the task target.

\begin{figure}[ht]
    \centering
    \includegraphics[width=0.9\textwidth]{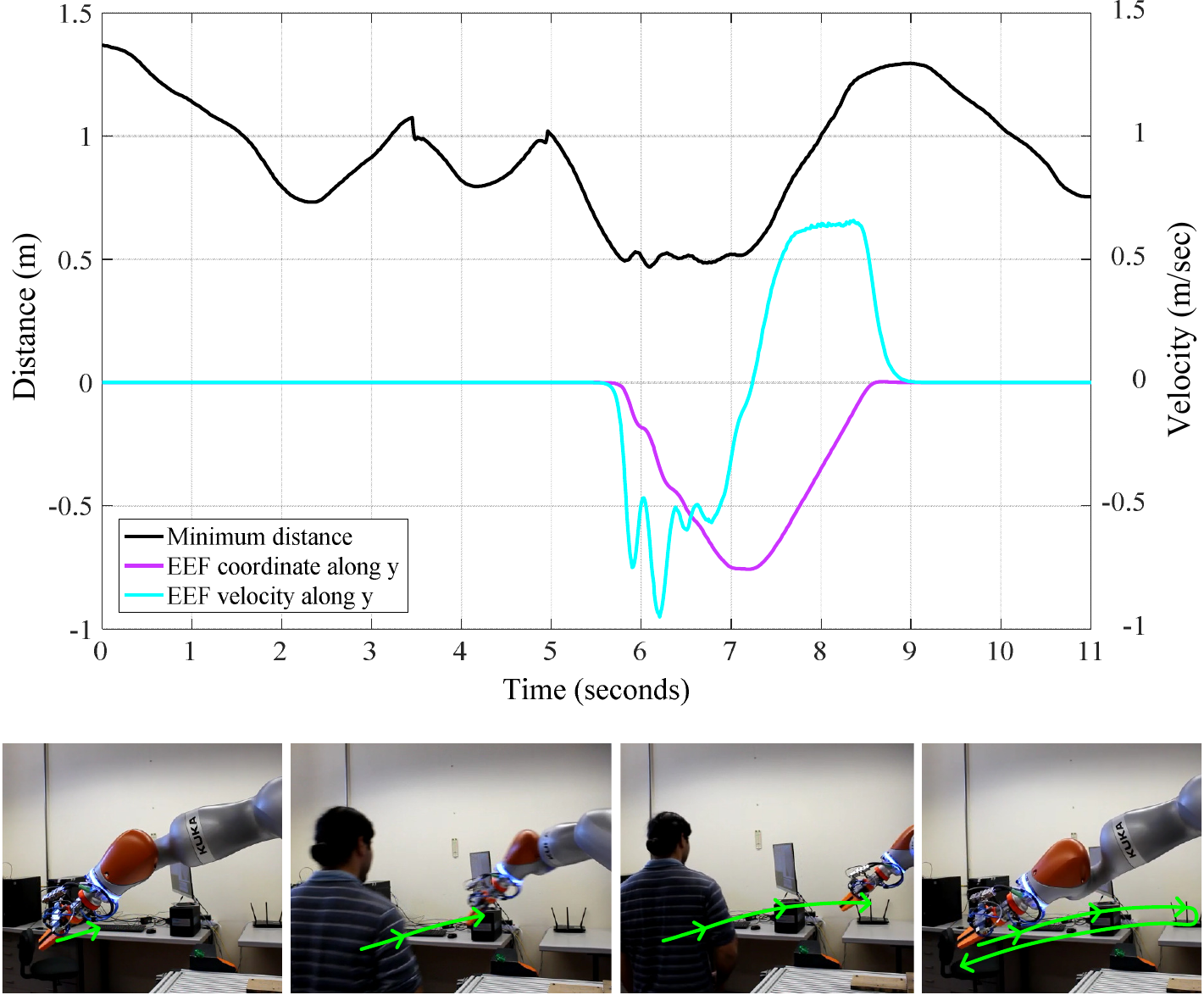}
    \caption{Configuration 2 results. Minimum distance, EEF velocity and position
along y axis (top). Snapshot of collision avoidance testing (bottom).}
    \label{fig:fig9}
\end{figure}

In configuration 2 the human approaches the robot from the side while
the robot is stopped at a pre-defined home position. As the human
approaches the robot the human-robot minimum distance decreases and
the robot reacts in an agile-smooth behaviour to avoid collision.
At the top of Fig. \ref{fig:fig9}, the same quantities
presented for configuration 1 show that at the beginning the human
starts walking towards the robot, when the minimum distance reaches
0.5 meters, the robot reacts to avoid collision. When the human goes
away the robot returns to the initial home position. The robot successfully
avoided collision as in the snapshots at the bottom of Fig. \ref{fig:fig9}
and in the video in extra multimedia materials.

\begin{figure}[ht]
    \centering
    \includegraphics[width=\textwidth]{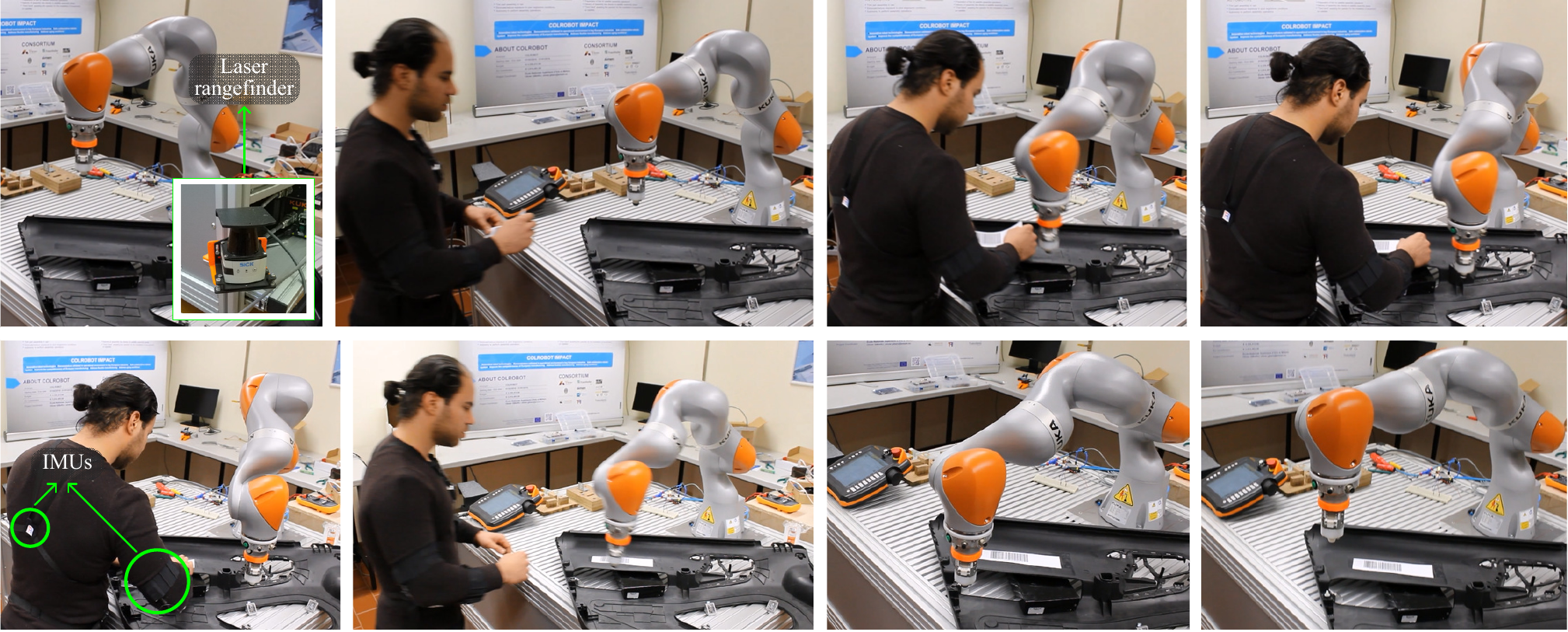}
    \caption{Automotive sample case for car door card assembly (video in extra
multimedia materials).}
    \label{fig:fig10}
\end{figure}

\begin{figure}[ht]
    \centering
    \includegraphics[width=\textwidth]{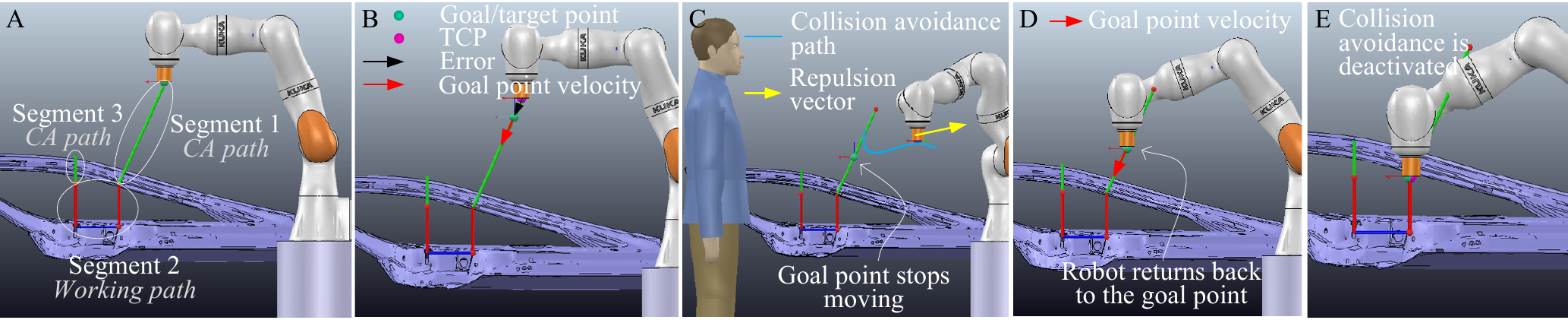}
    \caption{Pre-established sub-path segments for the automotive sample case.
This process is detailed in the algorithm Collision avoidance - automotive
sample case.}
    \label{fig:fig11}
\end{figure}

Flexible manufacturing, and industrial assembly processes in particular,
present several challenges due to the unstructured nature of the industrial
environment. Some tasks are more suited to be executed by humans,
others by robots, and others by the collaborative work between human
and robot. The ability to have humans and robots working side-by-side
will bring enormous efficiency benefits to flexible manufacturing.
However, this scenario is challenging, due to the requirement of having
the robot avoiding collisions with the co-worker in real-time, allowing
him/her to focus on the manufacturing tasks and not on the potential
danger from the robot side. In this context, we tested the proposed
system, in configuration 3, in an automotive sample case in which
the human co-worker approaches the robot to place a sticker in a car
door card while the robot is inserting trim clips in the same door
card, Fig. \ref{fig:fig10} (video in extra multimedia materials).
This flexible collaborative task allows the co-worker to manage his/her
working time and sequencing of operations since he/she is free to
place the sticker in the door card at any time and devotes attention
to other tasks that he/she has to take care of. Meanwhile the robot
continues inserting the trim clips in the door card by using its force
feedback to compensate for deviations in the door card positioning.
When the human co-worker approaches the robot it adapts the nominal
path to avoid collisions in a smooth way while keeping its task. For
this sample case, the pre-established nominal path is divided in 3
sub-path segments, Fig. \ref{fig:fig11} (A). In segments
1 and 3 (green lines) the collision avoidance control is activated
(collision avoidance (CA) paths) while in segment 2 (red and blue
lines) is deactivated. This is because this path is defined to be
the working path where the robot is inserting the trim clips at relative
reduced velocity. Starting from a given home position coincident with
the beginning of segment 1 and the tool centre point (TCP) the system
behaves as in Algorithm “collision avoidance - automotive sample case”.
In Fig. \ref{fig:fig11} (B) the robot and goal point
move along segment 1 so that an error vector is established. If the
human is detected in the safety zone the collision avoidance is activated
and the goal point stops moving, Fig. \ref{fig:fig11}
(C). When the human is not in the safety zone the robot returns back
to the goal point that starts moving with the robot, Fig. \ref{fig:fig11}
(D). When the robot reaches segment 2 the collision avoidance is deactivated,
Fig. \ref{fig:fig11} (E)\textcolor{black}{. Due to
the use of IMUs to capture the configuration of the human, in practice,
the noise in the measurements shall be filtered. In our study, we
utilized the Technaid IMU system, where filters \cite{Nogueira2017}
are implemented on an external processing unit (inside the system’s
hub). Consequently, using the manufacturer’s API (Application Programming
Interface) we can acquire the orientation of each IMU from an external
PC, wirelessly through a Bluetooth connection. Since that the accuracy
of the IMU measurements is represented by a maximum orientation error
(for example, the Technaid data sheets specify a one-degree angular
error), then based on the kinematics of the human body \cite{Mohammad2018},
an estimate of the resulting maximum Cartesian error can be calculated.
In such a case, our algorithm allows minimizing the effect of the
Cartesian error (due to the inaccuracy of the IMU angular measurement)
simply by increasing $d_{cr}$ (equation \ref{eq:repulsion_force_mag})
to include the maximum Cartesian error. In such a case, the uncertainty
in the calculated minimum distance (resulting from IMU inaccuracy)
will be less than the critical distance $d_{cr}$ (the minimum distance
below which the robot cannot be near the human), consequently, we
can be sure that the robot does not touch the human even when a maximum
measurement error is present.}

\textcolor{black}{Based on the tests, all the users indicate that
the system does not appear to be dangerous virtue to the collision
avoidance motion which is perceived as smooth and natural. It is also
demonstrated that the system performs well even in situations where
the human is showing a hesitation or a back-and-forth motion, this
is shown during tests (in the attached video segment, from seconds
43 to 47), where the co-worker is moving his hand back and forth towards
the robot while the robot is reacting to avoid collisions smoothly.
However, a final determination of the effect on the co-worker’s psychology
(feelings of danger, fear, security, distraction) will require a dedicated
study in collaboration with psychologists, involving more users and
taking into consideration various factors (including: age and background),
as such it will be left open for future work.}

\textcolor{black}{It is hard to perform a quantitative comparison
between our algorithm and others, since that various studies have
utilized different hardware architectures (robots, control systems),
different type of sensors, and different scenarios. However, qualitatively
speaking, our study, unlike others, demonstrates the feasibility of
on-line collision avoidance in HRI situation for a real industrial
use case while taking into consideration various important aspects:}
\begin{enumerate}
\item \textcolor{black}{Unlike other methods, our method is tailored for
industrial application, our algorithm implements off-line generated
paths of the EEF (important for performing industrial operation) combined
with an on-line reactive collision avoidance, to avoid collisions
with the human co-worker;}
\item \textcolor{black}{The algorithm takes into consideration various aspects
to guarantee the continuity of the motion (due to the necessity of
switching between different controllers imposed by the industrial
task);}
\item \textcolor{black}{In our tests, it is used a real industrial manipulator
KUKA iiwa (certified for industrial use), and utilizing commonly-used
sensors (well-established technologies);}
\item \textcolor{black}{A unique feature of our application is that we developed
the low-level control of our algorithm using the DirectServo library
of the KUKA iiwa. In such a case, all the security features of the
Sunrise.OS (operating system of the KUKA iiwa controller) are running
in parallel with our developed controlling program. This includes
the collision detection feature, which is activated by our control
program during the free collision avoidance motion, adding an additional
safety layer to the system;}
\item \textcolor{black}{Unlike most of the studies, we demonstrated our
algorithm in the context of a real-life (not simulation) industrial
scenario (assembly operation).}\floatname{algorithm}{Algorithm 2}
\end{enumerate}

\begin{algorithm}
\caption{Collision avoidance - automotive sample case}
\begin{algorithmic}[1]
\For{\textbf{each} time step}
\If{human is not detected in the safety zone}
\If{CA path}
	\State Goal point moves along path segment
	\State Error vector is generated between TCP and the goal, Fig. \ref{fig:fig11} (B)
	\State Attraction velocity is generated from the error vector
\If{goal point reaches the segment end}
	\State Goal point stops moving
	\State The robot TCP reaches the goal point
\EndIf
\Else
	\State Collision avoidance controller is deactivated
	\State Controller to insert trim clip is activated
\EndIf
\Else { human is detected in the safety zone}
\If{CA path}
	\State Goal point stops moving
	\State Integral term stops accumulating
	\State A repulsion velocity $v_{rep}$ acts in the robot, Fig. \ref{fig:fig11} (C)
\Else
	\State Collision avoidance controller is deactivated
	\State Controller to insert trim clip is activated, Fig. \ref{fig:fig11} (E)
\EndIf
\EndIf

\EndFor
\end{algorithmic}
\end{algorithm}

\section{Conclusion \label{sec:Conclusion-and-Future}}

This paper presented a novel method for human-robot collision avoidance
for collaborative robotics tailored for industrial applications. The
collision avoidance controller demonstrated on-line capabilities to
avoid collisions while the robot continues working by keeping the
task target. The concept of repulsion-vector-reshaping was introduced
to guarantee the continuity of the generated motion when switching
between controllers. Experiments indicated that the robot reaction
to avoid collisions is well perceived by the co-worker, smooth, natural
and effective.

\section{Acknowledgements}

This research was partially supported by Portugal 2020 project DM4Manufacturing
POCI-01-0145-FEDER-016418 by UE/FEDER through the program COMPETE
2020, and the Portuguese Foundation for Science and Technology (FCT)
SFRH/BD/131091/2017 and COBOTIS (PTDC/EMEEME/ 32595/2017).

\bibliography{library}

\bibliographystyle{abbrv}

\end{document}